\documentclass[10pt, a4paper]{article}

\usepackage[final]{lrec2026} 
\usepackage{tabularx}
\usepackage{multirow}
\usepackage{makecell}
\usepackage{subcaption}
\usepackage{float}

\title{A multilingual hallucination benchmark: MultiWikiQHalluA}

\name{Freja Thoresen, Dan Saattrup Smart}

\address{Alexandra Institute \\
         Rued Langgaards Vej 7, 2300 København \\
         freja.thoresen@alexandra.dk, dan.smart@alexandra.dk}

\abstract{
Most hallucination evaluations focus on English, leaving it unclear whether findings transfer to lower-resource languages. We investigate faithfulness hallucinations, defined as model-generated content that is fluent and plausible but diverges from the provided input or is internally inconsistent. 
Leveraging the multilingual MultiWikiQA dataset, we utilize the LettuceDetect framework to create synthetic hallucination datasets for 215 languages, from which we train token-level hallucination classifiers for 30 European languages. In this work, we present evaluations of model hallucinations on a selection of languages: English, Danish, German, and Icelandic. Using these classifiers, we evaluate the hallucination rates for Qwen3-0.6B, Qwen3-14B, Gemma-3-12B-IT, cogito-v1-preview-qwen-32B, and cogito-v1-preview-llama-70B.
Our classifiers reveal notably higher hallucination rates for Qwen3-0.6B (up to 60\% of answers containing at least one hallucination, peaking in Icelandic) and generally lower rates for larger models, with cogito-v1-preview-qwen-32B and cogito-v1-preview-llama-70B performing best on most languages. Hallucination rates are consistently higher for lower-resource languages, particularly Icelandic.
\\ \newline \Keywords{hallucination detection, multilingual natural language processing, token-level classification} 
}

\begin{document}

\maketitleabstract

\section{Introduction}
Large Language Models (LLMs) are prone to generating fluent yet false outputs, which is known as hallucinations.
We adopt the definition of faithfulness hallucinations as proposed by \citet{huang2025hallucination}: a language model generates fluent and plausible content that diverges from the given input/prompt, or is internally inconsistent. For example, if a model is asked to summarise a passage about climate change and introduces a claim not present in the source text. This is distinct from factuality hallucinations, which involve factual errors with respect to real-world knowledge regardless of what input was provided, for example, a model stating that the Eiffel Tower is located in London. Accordingly, the evaluation frameworks in this work focus on internally inconsistent or ungrounded model behaviour rather than external factual correctness.

Studies assessing language models’ factuality or evaluating whether the methods are effective to mitigate model hallucinations use different datasets and metrics. This makes it difficult to compare, in the same conditions, the factuality of different models as well as to compare the effectiveness of hallucination detection approaches. In this work, we use the same dataset, the open multilingual MultiWikiQA dataset by \citetlanguageresource{smart2025}, to evaluate models in the different languages. 

Most hallucination evaluations are conducted in English, leaving it unclear whether findings transfer to lower-resource languages. English, German, Danish, and Icelandic span a spectrum from highly to minimally represented in LLM pretraining corpora, providing a natural setting to study how language resource availability affects hallucination behaviour. We release an open source synthetic hallucination dataset covering 215 languages and train token-level classifiers for 30 European languages; in this paper we report evaluation results for four of those languages (English, Danish, German, and Icelandic).

In summary, our contributions are:
\begin{itemize}
  \item Release a synthetic hallucination dataset for 215\footnote{Originally planned 306, but we ran short of resources. If you need an additional language, please use \href{https://github.com/alexandrainst/factuality_eval}{this repository} to generate hallucinations.} languages (covering the full language support of MultiWikiQA).
  \item Release token-level hallucination classifiers for 30 European languages (a subset of the dataset languages for which we fine-tune models).
  \item Evaluate hallucination rates for five language models on four languages (English, Danish, German, and Icelandic).
\end{itemize}

\section{Related Work}

Hallucinations in language model outputs are commonly categorised into two types: factuality and faithfulness \cite{huang2025hallucination}. Factuality hallucinations involve claims that contradict established world knowledge (e.g.\ stating that the Eiffel Tower is in London). Faithfulness hallucinations occur when generated text diverges from a provided source context, such as introducing unsupported claims when summarising a passage.

Factuality benchmarks assess a model's parametric knowledge. FEVER \cite{thorne2018fever} verifies claims against evidence corpora; FActScore \cite{min2023factscore} evaluates atomic factual precision in long-form generations; TruthfulQA \cite{lin2022truthfulqa} probes susceptibility to common misconceptions; HaluEval \cite{li2023halueval} benchmarks hallucination detection across QA, summarisation, and dialogue; HalluLens \cite{bang2025hallulens} provides a broad multi-task evaluation of LLM hallucinations; and SimpleQA \cite{wei2024shortform} measures short-form factual accuracy. These approaches primarily test world knowledge and may miss context-grounded errors.

Faithfulness evaluation targets settings where generation should be grounded in a provided context, such as reading comprehension or Retrieval-Augmented Generation (RAG). NLI-based methods recast faithfulness verification as textual entailment: TRUE \cite{honovich2022true} shows that off-the-shelf NLI classifiers can serve as strong factual-consistency detectors. Other approaches include similarity-based metrics such as BERTScore \cite{zhang2019bertscore}, model-based judges such as Halu-J \cite{wang2024halu}, and stochastic self-consistency methods such as SelfCheckGPT \cite{manakul2023selfcheckgptzeroresourceblackboxhallucination}. Diagnostic frameworks such as RAGChecker \cite{ru2024ragchecker} further motivate evaluation beyond coarse answer-level correctness. Most recently, LettuceDetect \cite{kovacs2025lettucedetect} moves from answer-level to token-level hallucination detection, enabling precise localisation of unfaithful spans.

Notably, the benchmarks and detection methods described above focus predominantly on English, leaving it unclear whether findings transfer to lower-resource languages. We adopt the LettuceDetect approach for its token-level precision and extend it to a multilingual QA setting using MultiWikiQA \citelanguageresource{smart2025}, training hallucination detection models for 30 European languages spanning a range of resource levels.

\section{Methods}\label{sec:methods}
\begin{figure}[t]
  \centering
    \includegraphics[width=\columnwidth]{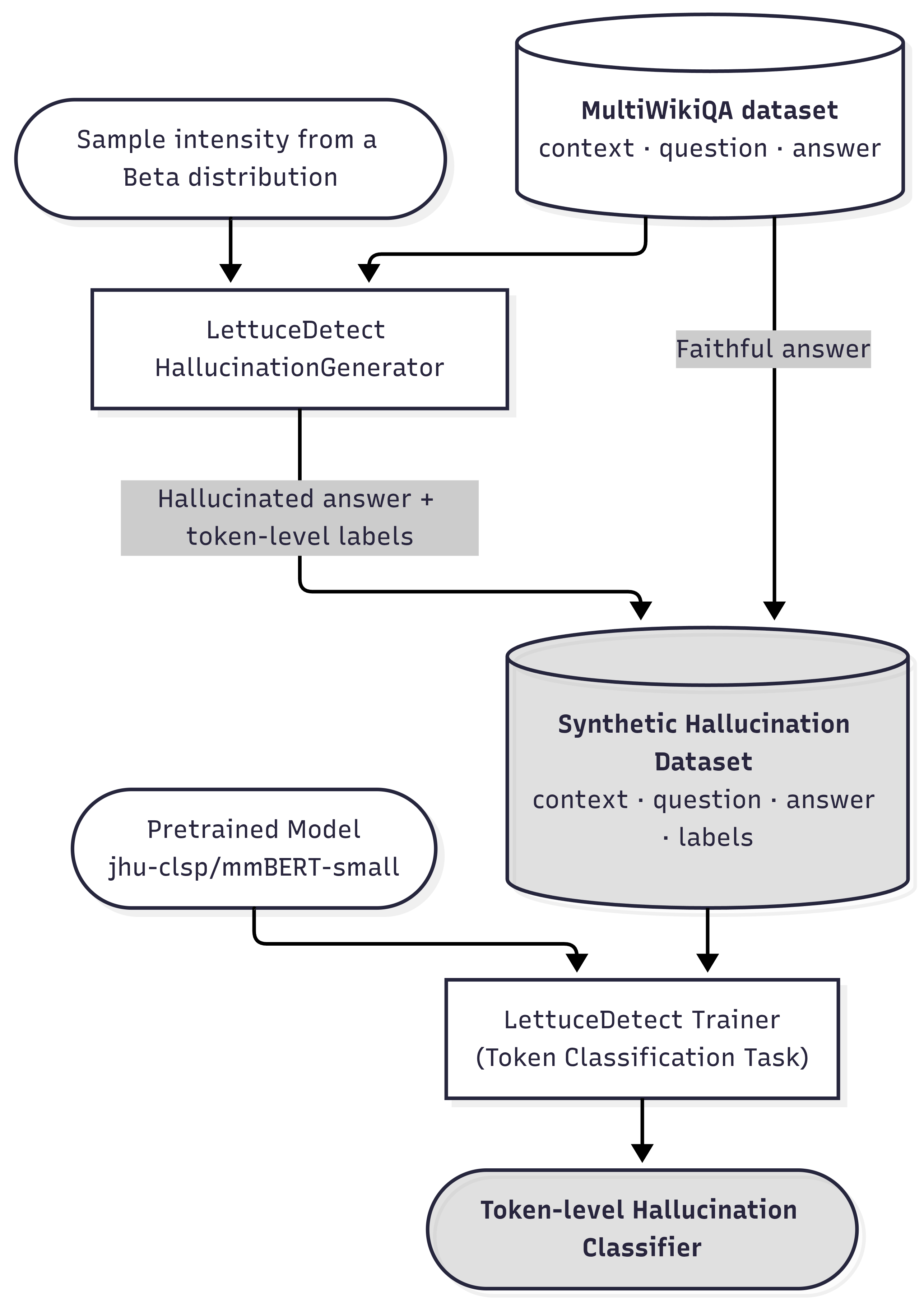}
    \caption{Overview of the two-stage methodology: synthetic hallucination data generation pipeline, where MultiWikiQA contexts, questions, and ground-truth answers are passed to the LettuceDetect framework, which uses a language model to produce token-labelled hallucinated answers; and fine-tuning of the mmBERT-small token-level hallucination classifier on the resulting dataset. The grey highlights the two deliverables: The synthetic hallucination dataset, and the token-level classifiers.}
    \label{fig:methods-overview}
\end{figure}

LettuceDetect \cite{kovacs2025lettucedetect} is a tool for detecting hallucinations in Retrieval-Augmented Generation (RAG) systems. It generates a hallucination dataset based on the dataset RagTruth \citelanguageresource{niu2024} and then trains a binary token-level classifier on it. This trained model can then be used to detect hallucinations in LLM-generated text in a reading comprehension context. LettuceDetect has multilingual support (7 languages) using EuroBERT from \cite{boizard2025} and implementations with small Ettin models from \cite{weller2025seqvsseq}. As a new addition to EuroBERT and Ettin models, we also train the mmBERT model from \cite{marone2025mmbert}, and we introduce two new languages (Icelandic and Danish) which were not previously supported by LettuceDetect.

\subsection{Datasets}
We use the open multilingual dataset MultiWikiQA \citelanguageresource{smart2025} as the foundation for all subsequent steps. The MultiWikiQA dataset supports 306 languages and contains context from Wikipedia articles, with questions, where the answers appear verbatim in the Wikipedia articles. In this study we evaluate on English, Danish, German, and Icelandic, while releasing resources for a broader set of 30 European languages. The training split includes 4000 context-question-answer triples and the test split contains 1000. In all subsequent experiments, models are evaluated on the exact same test set.

\subsection{Hallucination data generation}
The data generation works by supplying a dataset with context, questions, and answers, and the LettuceDetect framework will then generate hallucinated answers using a language model. Instead of using the RagTruth dataset, as originally used in the \cite{kovacs2025lettucedetect}, we use the MultiWikiQA dataset. We provide the LettuceDetect framework with the contexts, questions and answers from MultiWikiQA, and the framework then creates a false but plausible answer for each question. Hence, the result is a dataset with hallucinated answers, which can be used to train a classifier.  

For the LettuceDetect framework to generate a hallucinated dataset, it needs the following inputs: 

\begin{itemize}
  \item Dataset consisting of context, question, ground truth answer 
  \item Hallucination intensity 
  \item Language model to generate the hallucinated answer 
\end{itemize}

We provide the MultiWikiQA dataset separately for each language, the hallucination intensity is drawn for each sample (context, question, answer) from a beta distribution with mean of 0.2 and standard deviation of 0.15, and we use GPT-5-mini from OpenAI \cite{openai2024gpt4omini} as the language model. The beta distribution parameters were chosen to produce a distribution of hallucination intensities skewed toward subtle errors. The LettuceDetect framework then uses RAGFactChecker from Ru et al. \cite{ru2024ragchecker} to generate the hallucinated answer. RAGFactChecker will generate hallucinated answers based on the following rules on the hallucination intensity: 

\begin{itemize}
  \item Intensity $\le$ 0.2: Very subtle errors that are hard to detect 
  \item Intensity $\le$ 0.4: Moderate errors that are noticeable but plausible 
  \item Intensity $\le$ 0.6: Clear errors that are obviously incorrect 
  \item Intensity $\le$ 0.8: Strong errors that significantly change meaning 
  \item Intensity $>$ 0.8: Extreme errors that completely contradict the original 
\end{itemize}

RAGFactChecker can also create hallucinations on different error types. We use the default error types in LettuceDetect (and RAGFactChecker), which are the following: 

\begin{itemize}
\item Factual: Change specific facts, entities, or claims. 
\item Temporal: Modify dates, time periods, or temporal relationships. 
\item Numerical: Alter numbers, quantities, percentages or measurements. 
\end{itemize}
Concretely, RAGFactChecker instructs a language model to rewrite the reference answer according to the sampled intensity and error types, and to return (i) the rewritten answer text and (ii) a list of character-span pairs $[(s_1, e_1), (s_2, e_2), \ldots]$ marking every modified portion. These span annotations are projected onto the answer's subword tokens: any token whose character range overlaps with at least one hallucinated span is labelled 1 (\textit{unsupported}); all remaining tokens are labelled 0 (\textit{supported}). The generation results in a dataset with 5000 samples (4000 for training and 1000 for testing) with entirely hallucinated answers, for each language. 

\subsection{Classifier Training}
For each language, we use both the MultiWikiQA dataset with correct answers, and the hallucinated answer dataset generated with LettuceDetect. Hence, for each sample there is a "true" sample and a "hallucinated" sample, and both samples are used for training purposes, with binary labels assigned per token. To select the best base model for our classifier, we finetuned the token-level classifiers on the models in Table~\ref{tab:classifier-scores} using Danish and German. We chose these two languages because German is a high-resource language and Danish is a lower-resource language, allowing us to assess model performance across different resource levels. The F1-scores and accuracies are reported in Table~\ref{tab:classifier-scores}. The mmBERT-small model performed best in both Danish and German, and therefore we use the mmBERT-small model as the model to finetune for hallucination detection for European languages.

\begin{table*}[t]
\centering
\begin{tabularx}{\textwidth}{|X|X|X|X|X|}
\hline
\textbf{Model} & \textbf{Language} & \textbf{Supported-F1} & \textbf{Unsupported-F1} & \textbf{Accuracy} \\
\hline
Ettin-17m & Danish & 0.8239 & 0.6560 & 0.7670 \\
\hline
EuroBERT-210m & Danish & 0.9062 & 0.8206 & 0.8768 \\
\hline
mmBERT-small (140m) & Danish & \textbf{0.9143} & \textbf{0.8689} & \textbf{0.8963} \\
\hline
Ettin-17m & German & 0.8761 & 0.7291 & 0.8299 \\
\hline
EuroBERT-210m & German & 0.7737 & 0.4759 & 0.6839 \\
\hline
mmBERT-small (140m) & German & \textbf{0.9147} & \textbf{0.8627} & \textbf{0.8948} \\
\hline
\end{tabularx}
\caption{Classifiers finetuned with LettuceDetect on the MultiWikiQA train dataset with 4000 samples. The F1-scores and accuracies were evaluated from the test dataset with 1000 samples. The mmBERT-small model performed best in both Danish and German, and therefore we use the mmBERT-small model as the model to finetune for hallucination detection for European languages}
\label{tab:classifier-scores}
\end{table*}

\subsection{Model Evaluation}
When evaluating the models, we run model inference on the test set from the MultiWikiQA dataset. Then, we classify with the mmBERT-small finetuned classifier for each token if it was hallucinated or not. We evaluate Qwen3-0.6B and Qwen3-14B from Yang et al.\ \cite{yang2025qwen3}, Gemma-3-12B-IT \cite{team2025gemma3}, cogito-v1-preview-qwen-32B and cogito-v1-preview-llama-70B \cite{deepcogito2025cogito}. The results are presented in Table~\ref{tab:hallucination-scores}.

\begin{table*}[t]
\centering
\setlength{\tabcolsep}{4pt}
\begin{tabular}{|l|l|ccccc|}
\hline
\textbf{Metric} & \textbf{Language} & \textbf{Qwen3-0.6B} & \textbf{Qwen3-14B} & \makecell{\textbf{Gemma-3}\\\textbf{-12B-IT}} & \makecell{\textbf{Cogito-Qwen}\\\textbf{-32B}} & \makecell{\textbf{Cogito-Llama}\\\textbf{-70B}} \\
\hline
\multirow{4}{*}{\makecell[l]{Hallucination\\rate}} 
  & DA & 0.17 & 0.08 & 0.08 & \textbf{0.07} & \textbf{0.07} \\
  & DE & 0.09 & \textbf{0.03} & 0.05 & 0.05 & 0.05 \\
  & EN & 0.03 & \textbf{0.01} & 0.02 & \textbf{0.01} & 0.02 \\
  & IS & 0.36 & 0.17 & 0.20 & 0.18 & \textbf{0.15} \\
\hline
\multirow{4}{*}{\makecell[l]{Answer-level\\rate}} 
  & DA & 0.52 & 0.12 & 0.13 & 0.09 & \textbf{0.08} \\
  & DE & 0.17 & \textbf{0.04} & 0.06 & 0.06 & 0.06 \\
  & EN & 0.07 & 0.02 & 0.03 & \textbf{0.01} & 0.03 \\
  & IS & 0.60 & 0.26 & 0.27 & \textbf{0.18} & 0.19 \\
\hline
\end{tabular}
\caption{Hallucination scores by the finetuned mmBERT-small classifier for four languages: English (EN), Danish (DA), German (DE), and Icelandic (IS). \emph{Hallucination rate} is the token-level rate (hallucinated tokens / total tokens); \emph{Answer-level rate} is the fraction of answers containing at least one hallucinated token. Bold indicates the best (lowest) score per language per metric.}
\label{tab:hallucination-scores}
\end{table*}

\section{Discussion}


Across all models, high-resource languages (English and German) exhibit consistently lower hallucination rates than the lower-resource languages Danish and Icelandic, with Icelandic showing the highest rates. For the high-resource languages, the token-level hallucination rate remains low across all models except the smallest, whereas Danish and especially Icelandic reach notably higher rates. This pattern is more pronounced in the answer-level metric: for Icelandic, up to 60\% of answers contain at least one hallucinated token with Qwen3-0.6B.

The two larger models, cogito-v1-preview-qwen-32B and cogito-v1-preview-llama-70B, achieve the lowest or tied-lowest hallucination rates on three of the four languages, while Qwen3-14B performs best on German. On Icelandic, cogito-v1-preview-llama-70B achieves the lowest token-level rate of 0.15, while cogito-v1-preview-qwen-32B achieves the lowest answer-level rate of 0.18. The smallest model, Qwen3-0.6B, shows substantially higher hallucination rates across all languages, with Icelandic being particularly affected.

Notably, the relationship between model size and hallucination rate is not strictly monotonic. For example, Qwen3-14B outperforms the larger cogito-v1-preview-qwen-32B on German, and cogito-v1-preview-llama-70B does not always outperform cogito-v1-preview-qwen-32B. This suggests that architecture, training data composition, and multilingual coverage may matter as much as raw parameter count for hallucination behaviour across languages, however a larger sample size is needed in order to draw conclusions.

The LettuceDetect approach proved practical for our multilingual setting. Although dataset generation and classifier training are one-time costs, inference-time hallucination scoring is fast, making the approach scalable for large-scale evaluation across many languages. However, the classifier may overestimate the hallucination rate due to false positives, particularly for lower-resource languages where training signal is noisier. Further experiments such as varying the hallucination intensity distribution or cross-validating against larger human annotation sets are needed to quantify this bias.

Another potential confound is tokenization: low-resource languages tend to produce more tokens per sentence than high-resource languages \cite{rust-etal-2021-good}, because subword tokenizers trained predominantly on high-resource data split unfamiliar words into smaller pieces. This means that, for the same semantic content, a low-resource language may present more tokens to the classifier, increasing the opportunity for hallucination labels and inflating token-level hallucination rates. Disentangling the effect of tokenization granularity from genuine hallucination behaviour is an important direction for future work.

\section{Conclusion}

In this work, we presented a multilingual hallucination benchmark leveraging the LettuceDetect framework and the MultiWikiQA dataset. We released a synthetic hallucination dataset for 215 languages and token-level hallucination classifiers for 30 European languages, and evaluated five language models (Qwen3-0.6B, Qwen3-14B, Gemma-3-12B-IT, cogito-v1-preview-qwen-32B, and cogito-v1-preview-llama-70B) on English, Danish, German, and Icelandic. Our finetuned mmBERT-small classifiers showed strong calibration on gold answers and revealed that hallucination rates are consistently higher for the lower-resource language Icelandic. Among the evaluated models, cogito-v1-preview-qwen-32B and cogito-v1-preview-llama-70B achieved the lowest hallucination rates on most languages, while Qwen3-14B performed best on German. Model size alone did not determine hallucination behaviour, suggesting that architecture and multilingual training data composition play an important role.

\section{Resources}
All resources are publicly available. Note that the \textbf{dataset} covers 215 languages, the \textbf{classifiers} are released for 30 European languages (the subset for which we fine-tuned models), and the \textbf{evaluations} in this paper cover four languages (English, Danish, German, and Icelandic).
\begin{itemize}
  \item \textbf{Dataset}: The synthetic hallucination dataset for 215 languages is available on \href{https://huggingface.co/datasets/alexandrainst/multi-wiki-qa-synthetic-hallucinations}{HuggingFace}.
  \item \textbf{Models}: The finetuned mmBERT-small hallucination classifiers for 30 European languages are available as a \href{https://huggingface.co/collections/alexandrainst/mmbert-small-hallucination-token-classifiers}{HuggingFace model collection}.
  \item \textbf{Code}: The code for data generation, training, and evaluation is available on \href{https://github.com/alexandrainst/factuality_eval}{GitHub}.
\end{itemize}

\section{Authors’ Contributions}
\textbf{Freja Thoresen}: Conceptualization, Data Curation, Formal Analysis, Investigation, Methodology, Resources, Software, Validation, Visualization, Writing – Original Draft, Writing – Review \& Editing.

\textbf{Dan Saattrup Smart}: Conceptualization, Project Administration, Resources, Supervision, Writing – Review \& Editing.

\section{Acknowledgements}
This research was funded by the EU Horizon project
TrustLLM (grant agreement number 101135671).
\newline
\newline
\section{Bibliographical References}\label{sec:reference}

\bibliographystyle{lrec2026-natbib}
\bibliography{lrec2026-example}

\section{Language Resource References}
\label{lr:ref}
\bibliographystylelanguageresource{lrec2026-natbib}
\bibliographylanguageresource{languageresource}

\end{document}